\pdfoutput=1

\documentclass[11pt]{article}

\usepackage[review]{ACL2023}
\usepackage{amsmath, amssymb, wasysym}
\usepackage{times}
\usepackage{latexsym}
\usepackage{multirow}
\usepackage{graphicx}
\usepackage{verbatim}
\usepackage{float}
\usepackage[T1]{fontenc}

\usepackage[utf8]{inputenc}
\usepackage{comment}
\usepackage{microtype}

\usepackage{inconsolata}
\usepackage{tcolorbox} 
\tcbuselibrary{skins, breakable} 
\usepackage{xcolor} 
\usepackage{listings}
\title{PRiME: \underline{P}ragmatic \underline{R}easoning \underline{I}n \underline{ME}ntal health}

\author{Sneha Oram \\
  Affiliation / Address line 1 \\
  Affiliation / Address line 2 \\
  Affiliation / Address line 3 \\
  \texttt{email@domain} \\\And
  Second Author \\
  Affiliation / Address line 1 \\
  Affiliation / Address line 2 \\
  Affiliation / Address line 3 \\
  \texttt{email@domain} \\}

\begin{document}
\maketitle
\begin{abstract}
In this work, the task and domain interaction of reasoning and mental health are investigated. As large language models (LLMs) present exceptional text understanding and fluent text generation capability, we analyze whether they can reason in the high-stakes domain of mental health. We begin by introducing three datasets and three tasks leveraging the existing work in pragmatic reasoning. The analysis constitutes an evaluation of $12$ LLMs, including MentaLLaMA, which is a mental health reasoning and explanation generation model. The performance of the models remains on par with human evaluation, with the Mistral-7B model giving the best performance. We further analyze the performance of the models with error asymmetry ratio and conduct \textit{first-order attention rollout experiment} with MentaLLaMA-7B. We present the results for the community reporting the architectural behaviour of the models for the input text and recommend further careful analysis for the high-stakes domain.

\end{abstract}

\section{Introduction}
Artificial intelligence (AI) is increasingly being integrated into our everyday lives, driven by recent rapid advancements in the field. At the same time, mental health conditions continue to affect millions of individuals worldwide \cite{gbd2022global, patel2018lancet}, leading to people increasingly turning to AI for mental health support. At this stage, it is crucial to critically evaluate the progress of AI, particularly with natural language processing (NLP), and its implications for high-stakes domains, such as mental health. In mental health applications, automated systems such as chatbots must accurately interpret nuanced expressions and the implicit intent of human language. Motivated by this requirement, we investigate \texttt{pragmatics} in mental health by adopting large language models (LLMs). As this setting supports a discourse and conversational scenario, we consider examining pragmatics for our use case. In particular, we undertake pragmatic phenomena and develop complex tasks for mental health analysis, motivating from \textit{pragmatic reasoning tasks} in NLP. Following this, we conduct a rigorous analysis with LLMs. Our evaluation of the pragmatic understanding of LLMs concerns the discussion of the deployment of automated systems in the domain. Our analysis of multiple LLMs shows that the models demonstrate performance which are at par with human judgement in our proposed tasks. 

\begin{figure*}[ht]
    \centering
    \includegraphics[scale=0.85]{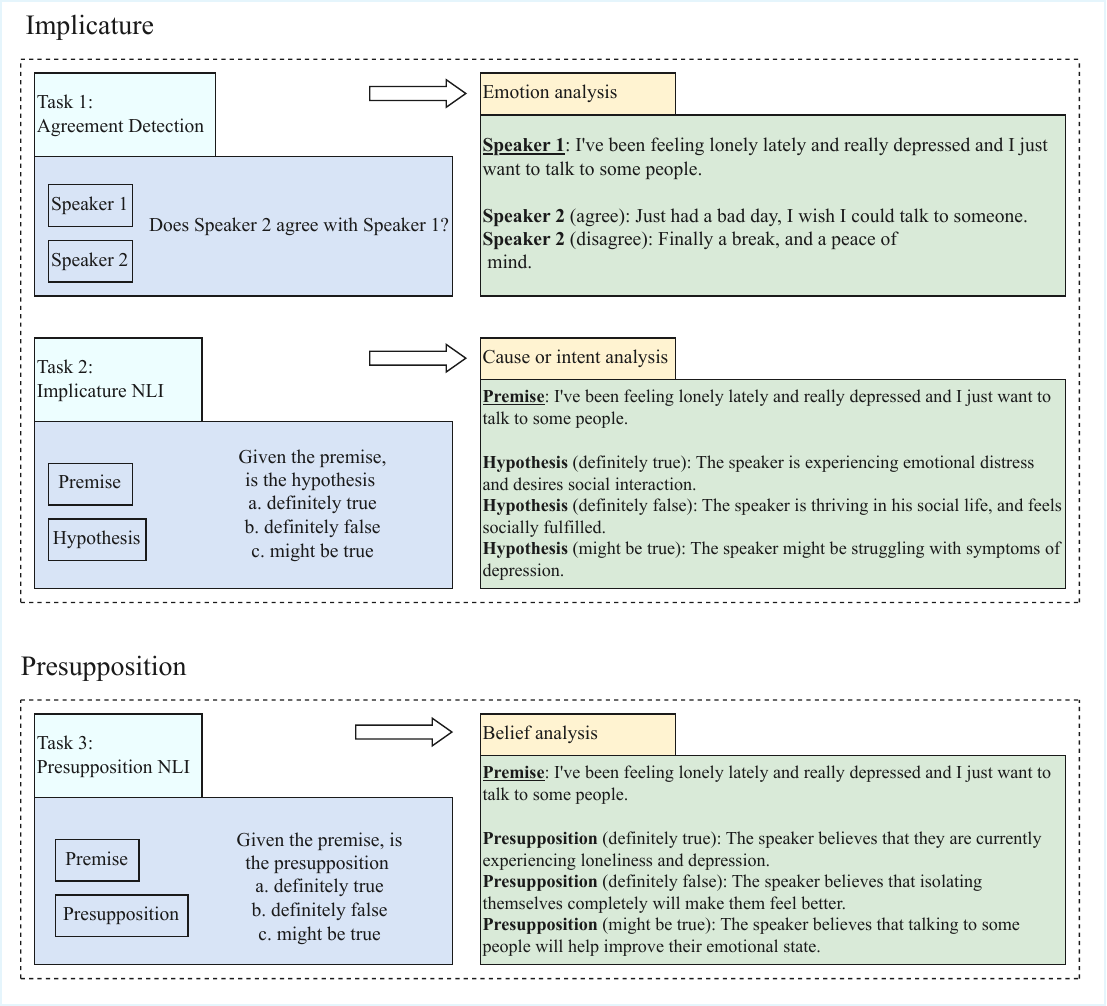} 
    \caption{Overview of our streamlined pragmatics reasoning tasks in mental health with examples.}
    \label{fig:frontpage}
\end{figure*}

Prior work by \citet{10.36227/techrxiv.21972974.v1} demonstrates an essential gap with computational intelligence and mental health, which can be bridged with explainable and interpretable AI. This largely corresponds to pragmatic reasoning in the mental health domain. While existing research has investigated mental health LLMs and pragmatics separately to some extent, the interaction between these two topics remains largely underexplored. Consider a typical textual mental health expression: \textit{“I’ve been feeling lonely lately and really depressed, and I just want to talk to some people"}, along with an underlying statement, \textit{“The speaker might be struggling with symptoms of depression"}, which may be true. Given the advanced text understanding capabilities of LLMs, we aim to investigate whether LLMs can capture the implied meanings and interpret such underlying statements in mental health–related texts. 

To address this, we first introduce three datasets, collectively named the \textbf{PRiME} (\textbf{P}ragmatic \textbf{R}easoning \textbf{i}n \textbf{ME}ntal health) datasets, which cover different aspects of mental health. Corresponding to the datasets, we subsequently present three tasks for pragmatic reasoning in mental health, analyzing the capabilities of LLMs in mental health by leveraging the pragmatic phenomena of \textit{implicature} and \textit{presupposition}. Our comprehensive analysis with $12$ LLMs, reveals that they exhibit average mental health reasoning capabilities, with smaller models representing low performance. We present a benchmark data resource and corresponding tasks and evaluate with a set of $12$ LLMs to demonstrate the current progress of NLP models for pragmatic reasoning in mental health.

To further articulate, we streamline two existing tasks in \texttt{implicature}, namely, Agreement Detection and \texttt{Implicature} natural language inference (NLI). For \texttt{presupposition}, we streamline the existing task of \texttt{Presupposition} NLI. The details of the tasks can be found in Figure \ref{fig:frontpage}. 

\noindent Our contributions are: 
\begin{itemize}
    \item \textbf{PRiME Benchmark}: Three novel datasets and the corresponding tasks for \underline{P}ragmatic \underline{R}easoning in \underline{ME}ntal health (PRiME). Each dataset contains approximately $1400$ data points created with a combination of real data (\textit{publicly available}, utilized from \citet{garg-etal-2022-cams}) and synthetic data. The dataset will be open-sourced to the community for further research. 
    \item \textbf{PRiME Evaluation}: A systematic evaluation of the \textbf{PRiME} benchmark with $12$ LLMs. The following insights are centered on the performance of LLMs with respect to model size and versatility in different tasks. Also, a discussion on the performance of MentaLLaMA, with a benchmark analysis using \textit{attention rollout} experiment. 
\end{itemize}

\section{Related works}  \label{sec:relwork}

\noindent \textbf{Preliminaries:} Pragmatic reasoning in NLP is extensively explored in prior research works \cite{sravanthi-etal-2024-pub, zheng2021grice, kim-etal-2023-qa, kabbara2022investigating}. Previous studies have explored pragmatic reasoning in open-domain settings; however, in the context of mental health, we emphasize the need to refine standard pragmatic reasoning tasks. Compared to language use in open-domain settings, texts expressing mental health concerns typically exhibit a lower emotional valence and a higher degree of negative sentiment. In this work, we explore two pragmatic phenomena in the light of mental health, \texttt{implicature}, and \texttt{presupposition}.

Following the Handbook of Pragmatics \citet{horn2004handbook} and \citet{grice1975logic}, \texttt{implicature} indicates understanding what is suggested or implied in a statement which is not literally expressed. The \texttt{presupposition} reflects an implicit assumption that is taken for granted before the use of a statement. In the context of mental health, we propose a revision and streamlining of \texttt{implicature} to focus on emotion analysis and cause or intent analysis, and \texttt{presupposition} to focus on belief analysis. Specifically, we present revised definitions as:

\noindent \textbf{\texttt{Implicature}}: Understanding the emotion and implied cause or reason behind the speaker's feelings or emotional state, whether expressed implicitly or explicitly. 

\noindent\textbf{\texttt{Presupposition}}: Understanding the inherent assumption or belief of the speaker, to extract an underlying reason.

It may be noted that the \texttt{presupposition} operates at a deeper inferential level than the \texttt{implicature}, following the framework presented in the Handbook of Pragmatics \cite{horn2004handbook}.\footnote{It might be noted that in linguistics or psycholinguistics, the core concepts of pragmatics, implicature, presupposition might entail different interpretations. In our work, we mainly focus on the assessing reasoning capabilities of LLMs motivating from the standard NLP tasks and tie the use case of mental health.}

\subsection{LLMs in Mental Health}
LLMs have been widely adopted in prior research for developing chatbots that generate personalized empathetic responses to support mental health interventions \cite{hua2025large, saha2022mental, saha2022towards, mishra2023therapist, Mishra_Priya_Ekbal_2023}. Recent work on reasoning in mental health has explored alignment-based frameworks using reinforcement learning to improve LLM reasoning abilities \cite{dai2025psyche, feng2025stimuli, zeng2025psychcounsel, xiao2025mentrasuite}. However, such approaches incur high computational costs and are prone to catastrophic forgetting, where models lose old skills when learning new ones \cite{wolczyk2024fine, wolczyk2023role, schmied2023learning}, leading to degraded performance and, in our use case, a misclassification during inference.

\subsection{Pragmatics reasoning with LLMs}
Reasoning in NLP has been extensively studied in open-domain settings, including commonsense reasoning, mathematical problem solving, and reasoning over world knowledge \cite{sravanthi-etal-2024-pub, zheng2021grice, kim-etal-2023-qa, kabbara2022investigating}. In parallel, pragmatic reasoning has been explored through computational models that examine how humans use and interpret language in context \cite{grice1975logic}. However, \citet{10.36227/techrxiv.21972974.v1} identify a clear gap between research in pragmatics and its application to mental health contexts, which this work aims to address.    

\subsection{MentaLLaMA model}
In the above context, MentaLLaMA \cite{yang2024mentallama} represents a notable work to incorporate reasoning and interpretability into mental health–focused language models. It is fine-tuned on the interpretable mental health instruction (IMHI) dataset using two variants of LLaMA-$2$ \emph{i.e.,} LLaMA-$2$-$7$B-chat and LLaMA-$2$-$13$B-chat \cite{touvron2023llama2openfoundation} to generate explanatory responses. The model demonstrates the ability to identify underlying factors associated with mental health conditions and to produce coherent reasoning-based explanations. The IMHI dataset is constructed from social media posts, which encourages an evaluation of the performance of MentaLLaMA in our CAMS dataset as it reflects the data source and structure relative our CAMS dataset. 

Our proposed tasks harness pragmatic phenomena in linguistics and establish a framework for reasoning in mental health.


\section{Dataset}  \label{sec:dataset}
To facilitate the investigation of pragmatic reasoning in LLMs within the mental health domain, we introduce the \textbf{PRiME} dataset, which comprises approximately $3$x$1400$ data points. The dataset is constructed by combining the existing CAMS dataset \citep{garg-etal-2022-cams} with synthetic data generated using \texttt{GPT$5$}. The original CAMS dataset consists of Reddit posts collected from social media and spanning the period from $2015$ to $2018$.

\begin{table}[]
\centering
\begin{tabular}{|c|c|}
\hline
\textbf{Tasks}                                                          & \textbf{Label counts \& proportions}                                                                                           \\ \hline
\begin{tabular}[c]{@{}c@{}}Task 1:\\ Agreement\\ Detection\end{tabular} & \begin{tabular}[c]{@{}c@{}}\#Agree: $700$ ($50\%$)\\ \#Disagree: $700$  ($50\%$)\end{tabular}                                          \\ \hline
\begin{tabular}[c]{@{}c@{}}Task 2:\\ Implicature\\ NLI\end{tabular}     & \begin{tabular}[c]{@{}c@{}}\#Definitely true: $500$ ($35.78\%$)\\ \#Definitely false: $497$ ($35.57\%$)\\ \#Might be true: $400$ ($28.63\%$)\end{tabular} \\ \hline
\begin{tabular}[c]{@{}c@{}}Task 3:\\ Presupposition \\ NLI\end{tabular} & \begin{tabular}[c]{@{}c@{}}\#Definitely true: $497$ ($35.70\%$)\\ \#Definitely false: $494$ ($35.48\%$)\\ \#Might be true: $401$ ($28.80\%$)\end{tabular} \\ \hline
\end{tabular}
\caption{PRiME dataset statistics for three redefined tasks for implicature and presupposition in mental health.
}
\label{tab: 1}
\end{table}

The CAMS dataset is first cleaned and augmented with synthetic data. Subsequently, data validation and quality assessment are performed by human annotators, as discussed in the following sections. After this process, the final dataset is obtained, with summary statistics reported in Table~\ref{tab: 1}.

\subsection{Data cleaning} \label{data_cle}
Social media posts from Reddit contain substantial noise, including spelling and punctuation errors, non–mental health–related content, and the use of abbreviations. In many cases, the intended meaning of a post is also not clearly conveyed. To improve the suitability of the data for LLMs, we apply a series of cleaning steps: (i) moderating offensive language, (ii) expanding abbreviations to their full forms, (iii) removing inappropriate or disturbing content, and (iv) revising posts where the intended meaning is unclear. 

\begin{table}[t]
\centering
\begin{tabular}{c}
\hline
\textbf{Models}                                                                    \\ \hline
\begin{tabular}[c]{@{}c@{}}mistralai/Mistral-7B-Instruct-v0.2\\ \cite{jiang2023mistral}\end{tabular} \\ \hline
\begin{tabular}[c]{@{}c@{}}mistralai/Mistral-7B-Instruct-v0.3\\ \cite{jiang2024mistral}\end{tabular} \\ \hline
\begin{tabular}[c]{@{}c@{}}Qwen/Qwen3-4B-Instruct-2507\\ \cite{qwen3technicalreport}\end{tabular}        \\ \hline
\begin{tabular}[c]{@{}c@{}}Qwen/Qwen2.5-7B-Instruct\\ \cite{qwen2.5}\end{tabular}           \\ \hline
\begin{tabular}[c]{@{}c@{}}Qwen/Qwen2.5-1.5B-Instruct\\ \cite{qwen2.5}\end{tabular}         \\ \hline
\begin{tabular}[c]{@{}c@{}}google/gemma-2-2b-it\\ \cite{gemma_2024}\end{tabular}               \\ \hline
\begin{tabular}[c]{@{}c@{}}google/gemma-3-1b-it\\ \cite{gemma_2025}\end{tabular}               \\ \hline
\begin{tabular}[c]{@{}c@{}}meta-llama/Llama-2-7b-chat-hf\\ \cite{touvron2023llama2openfoundation}\end{tabular}      \\ \hline
\begin{tabular}[c]{@{}c@{}}meta-llama/llama-3.1-8B-Instruct\\ \cite{llama3_1_8b_instruct_2024} \end{tabular}   \\ \hline
\begin{tabular}[c]{@{}c@{}}meta-llama/Llama-3.2-1B-Instruct\\ \cite{llama3_2_1b_instruct_2024}\end{tabular}   \\ \hline
\begin{tabular}[c]{@{}c@{}}klyang/MentaLLaMA-chat-7B\\ \cite{yang2024mentallama}\end{tabular}          \\ \hline
\begin{tabular}[c]{@{}c@{}}klyang/MentaLLaMA-chat-13B\\ \cite{yang2024mentallama}\end{tabular}         \\ \hline
\end{tabular}
\caption{List of twelve LLMs used for evaluation on the PRiME dataset
}
\label{tab:3}
\end{table}

\subsection{Synthetic Data Generation}  \label{data_ge}
For the implicature study using agreement detection, social media posts from the original dataset are treated as utterances from Speaker $1$, while the corresponding responses from Speaker $2$ are synthetically generated using \texttt{GPT5}. Similarly, in Implicature NLI and Presupposition NLI, the posts serve as premises, and the corresponding hypotheses and presuppositions are generated synthetically. In terms of text composition, each of the full datasets contains 50\% human-authored content and 50\% synthetic content. 

\begin{table}[t]
    \centering
    \includegraphics[scale=0.6]{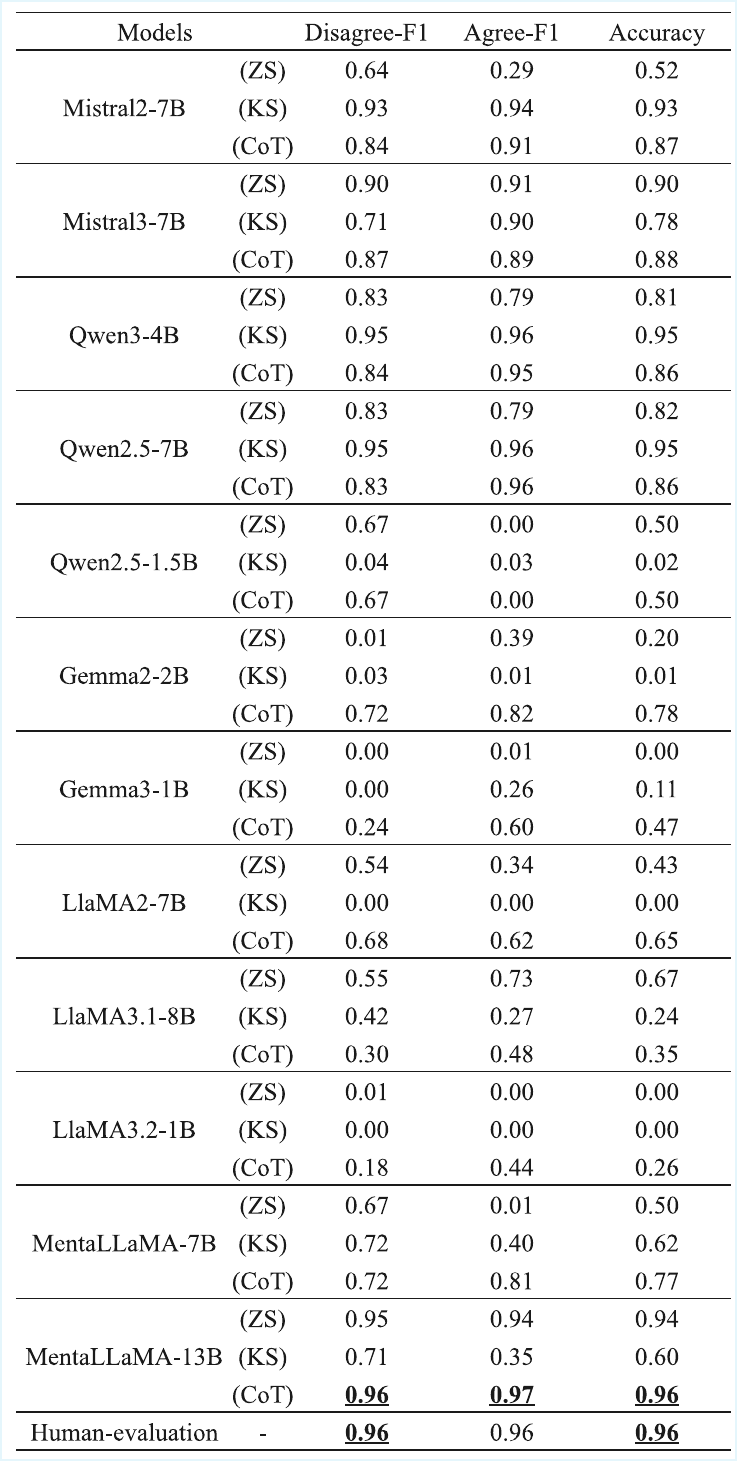}
    \caption{Performance of LLMs on the Agreement Detection task. ZS, KS, and CoT refer to zero-shot, k-shot, and chain-of-thought prompt settings, respectively.}
    \label{tab: AD}
\end{table}

\subsection{Data validation and quality check}  \label{data_chec}
Following data generation, the synthetic texts are manually evaluated and reviewed by human annotators. After this human filtering, we find no evidence of gender, socioeconomic, or racial bias in the synthetic data. However, the original CAMS dataset exhibits geographic and linguistic biases, as well as an age bias toward the teenage population in the United States. These biases arise from the nature of Reddit posts, which reflect specific user demographics. 

For the data annotation of NLI datasets, we worked with a practicing MD psychiatrist and a psychologist. The practitioners annotated the datasets following the guidelines that we prepared, and this is elaborated in the Appendix \S\ref{anno_gui}. After resolving annotation disagreements, the final ground truth labels obtained with inter-annotator agreement (IAA) scores of $\kappa = 0.72$ for Implicature NLI and $\kappa = 0.68$ for Presupposition NLI, respectively. The dataset for Agreement Detection is annotated by two retired government school teachers, who are qualified with a Bachelor's in Education (B.Ed.). The ground truth labels in the Agreement Detection task are obtained with IAA of $\kappa = 0.79$. With these IAA scores, the \textbf{PRiME} dataset is finalized as the gold standard dataset. The agreement score shows that the proposed dataset is a challenging classification task. 

All the annotators in this study are fairly compensated for their assigned tasks. The final dataset, \textbf{PRiME} will be made open-source for future research.

\section{Evaluation Methodology} \label{sec: method}
To evaluate the reasoning capabilities of LLMs in the mental health domain, we adopt a multiple-choice question answering (MCQA) prompting framework. Each input consists of a question paired with a set of candidate answers, each labeled with a symbolic identifier (\emph{e.g.,} A, B, C). Under this setting, we conduct experiments using zero-shot, $k$-shot, and chain-of-thought prompting strategies. For the $k$-shot prompting, we set $k=3$ for NLI tasks and $k=2$ for the Agreement Detection task to maintain the balance over the labels. The $k$-shot examples included in the prompt templates are excluded from the evaluation set, and the remaining data points are used for testing, for a fair classification. 
For demonstration, we present the prompt used for evaluation for the Implicature NLI task in the zero-shot setting. 
\begin{lstlisting}
You are a mental health support assistant. Given a premise and a hypothesis, answer whether the hypothesis is A. definitely true, B. definitely false, or C. might be true given the premise. 
Premise: {premise-text}
Hypothesis: {hypothesis-text}
Choose the correct option: A. definitely true, B. definitely false, or C. might be true
Response: [A/B/C]
\end{lstlisting}

\noindent The prompts used for other tasks with all three prompting strategies are provided in Appendix \S\ref{sec: eval}.

We conduct an additional attention rollout analysis \cite{DBLP:conf/acl/AbnarZ20} using MentaLLaMA. Due to the attention sink phenomenon observed in standard attention rollout, where the start token \textit{<s>} receives the maximum attention score, we instead apply order-level attention following the approach of \citet{liang2025orderlevel}. Instead of computing the consecutive matrix multiplication across all layers, \citet{liang2025orderlevel} utilized the order-level decomposition of the attention rollout, expressed as: 

\begin{equation}
\small
\begin{split}
\hat{A} = I 
+ \sum_{i=1}^{N} A^{(i)} 
+ \sum_{1 \le i \le j \le N} A^{(j)} A^{(i)} \qquad + \cdots  \\
\cdots 
+ A^{(N)} A^{(N-1)} \cdots A^{(1)}
\end{split}
\end{equation}

Here, $A$ represents the mean of the attention matrices over the heads in a layer, and $N$ is the total number of layers in a model. The first term in the equation corresponds to the residual connection, the second term corresponds to the sum of the $A$'s across the layers, called the first-order attention. This is extended for $k$-th order attention, with $A$ in layer $k$ is consecutively multiplied with $A$'s of lower layers of $k$. This step is repeated until the $N$-th order attention with $A$ of the last layer is consecutively multiplied with $A$'s of all lower layers. As  \citet{liang2025orderlevel} has shown, the first-order attention depicts the commonality among the language models, we chose it to demonstrate the performance of MentaLLaMA.

\begin{table*}[t]
    \centering
    \begin{minipage}{0.49\textwidth}
        \centering
        \includegraphics[scale=0.57]{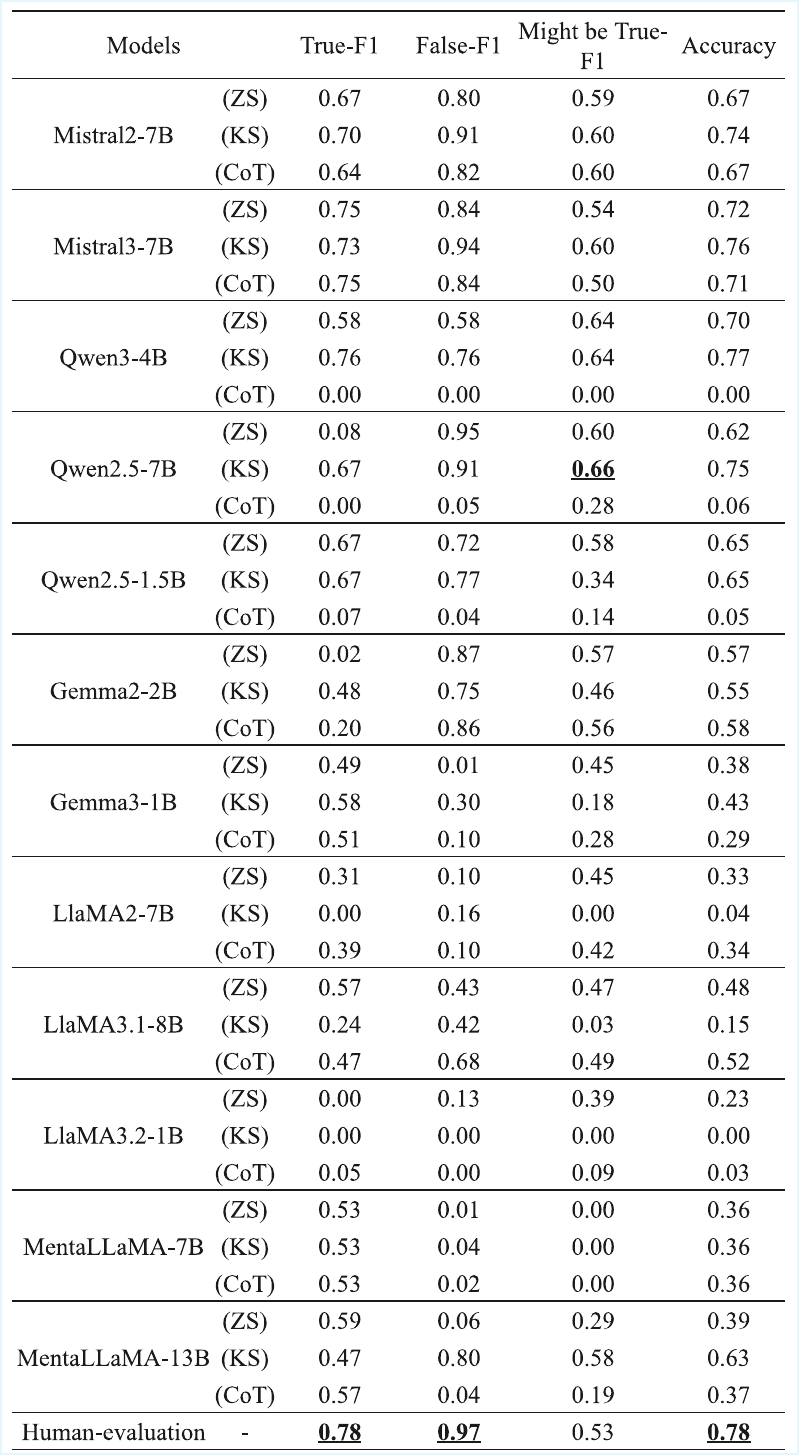}
        \caption{Performance of LLMs on the Implicature NLI task. ZS, KS, and CoT refer to zero-shot, k-shot, and chain-of-thought prompt settings, respectively. The label \lq definitely true\rq is interchanged with \lq true\rq, and \lq definitely false\rq is interchanged with \lq false\rq.}
        \label{tab: NLI}
    \end{minipage}
    \hfill
    \begin{minipage}{0.49\textwidth}
        \centering
        \includegraphics[scale=0.57]{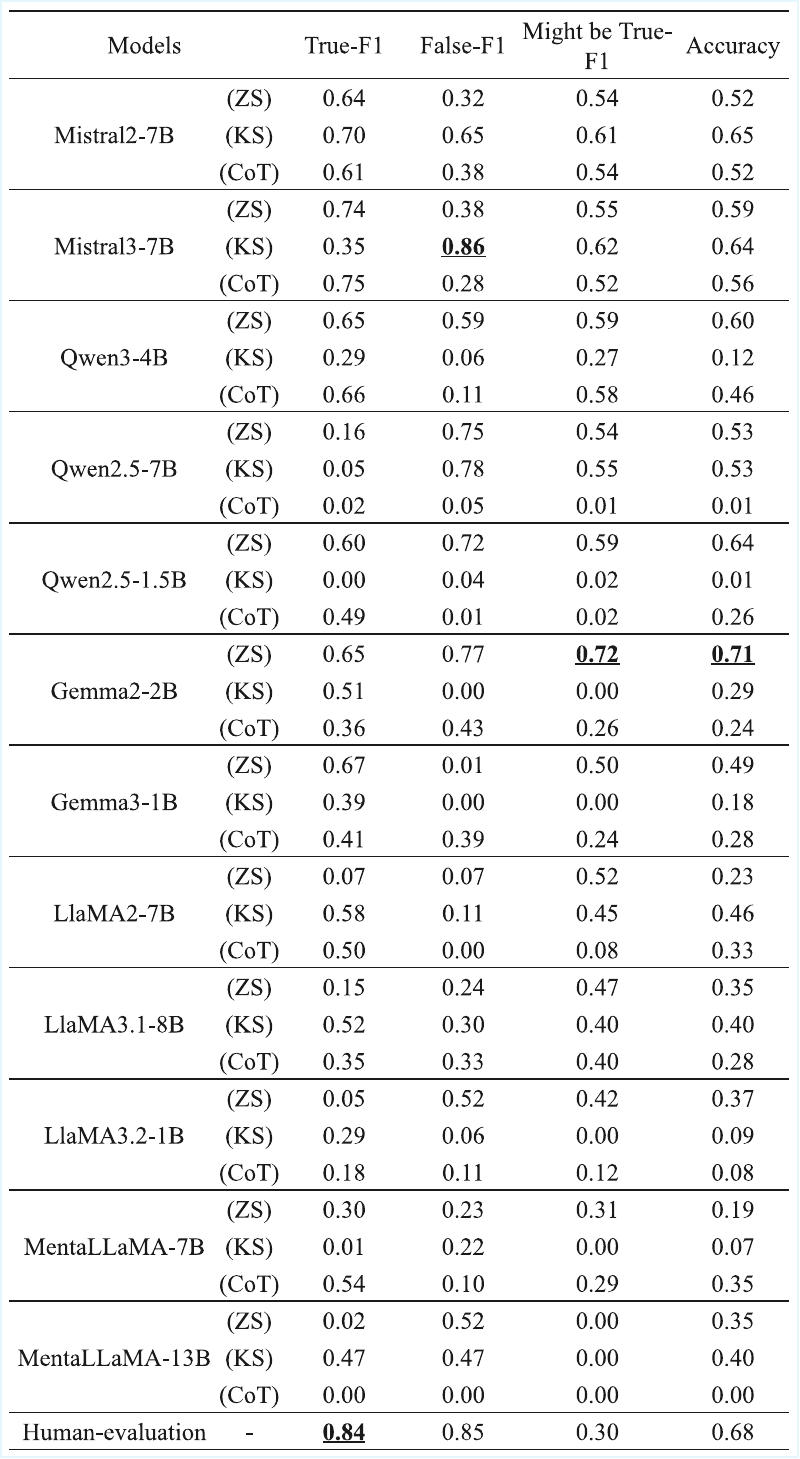}
        \caption{Performance of LLMs on the Presupposition NLI task. ZS, KS, and CoT refer to zero-shot, k-shot, and chain-of-thought prompt settings, respectively. The label \lq definitely true\rq is interchanged with \lq true\rq and \lq definitely false\rq is interchanged with \lq false\rq.}
        \label{tab: PNLI}
    \end{minipage}
\end{table*}

\section{Experiments}
The details of the $12$ instruction-tuned models used in the analysis are listed in Table \ref{tab:3}. The experiments are conducted with hyperparameter settings of top-p as $0.9$, temperature as $0.1$, and the sampling seed is $42$. Additionally, we chose the parameter of \texttt{max-new-tokens} as $8$ for zero-shot (ZS) and k-shot (KS) settings, and $100$ for the chain-of-thought (CoT) setting to allow the models to form reasoning chains. From the settings, the output of LLMs are found to be clean and unambiguous, consisting of a class label such as “agree”, “definitely true”, or “option A”. These predicted class labels are extracted by parsing the LLM outputs, which are subsequently compared with the true labels. For each task, we report the performance of all $12$ models under the three prompt settings. In total, the experimental setup comprises $12$ models, $3$ tasks, and $3$ prompt settings, resulting in $108$ experiments, and each of these experiment is repeated for $3$-$5$ runs. The evaluation is focused on classification with LLM-inferencing, and no training is conducted in the study. We utilized an NVIDIA A$100$ GPU for approximately $3.5$, $5$, and $6$ hours for the ZS, KS, and CoT prompting settings, respectively. We report class-wise F$1$-scores and the overall accuracy of the classification results. 

\section{Results and Analysis} \label{sec:results}

The performance of LLMs in the three tasks is presented in Tables \ref{tab: AD}, \ref{tab: NLI}, and \ref{tab: PNLI}. For comparison, we include a human evaluation baseline conducted by two Master’s students in Computer Science\footnote{who are compensated fairly, and obtained IAA ($\kappa = 0.76$). The instructions outlined in the Appendix \S \ref{sec: eval}.}. On the Agreement task, MentaLLaMA-$13$B with CoT prompting achieves the best performance, attaining an accuracy of $0.96$, which matches the human baseline. Overall, CoT and KS prompting significantly outperform ZS, particularly for smaller models such as Gemma$2$-$2$B. Larger models, such as Mistral and Qwen$3$-$4$B, demonstrate greater stability across settings, whereas smaller LlaMA variant (LLaMA$3.2$-$1$B) performs poorly, often yielding near-zero scores in certain categories. 

\begin{figure}[]
    \centering
    \includegraphics[width=1\linewidth]{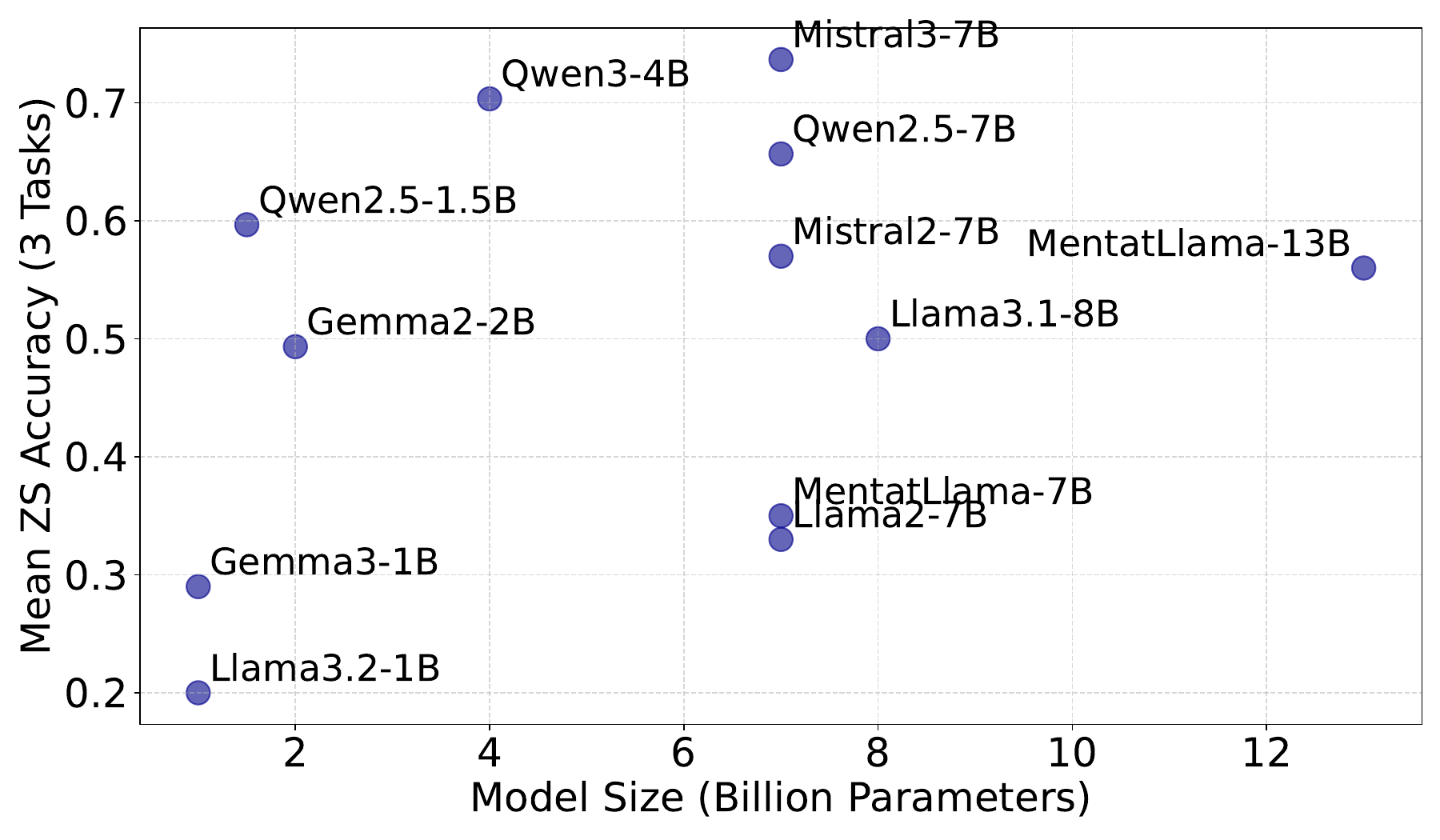} 
    \caption{Accuracy of the models with increasing model size on zero-shot (ZS) prompt setting.}
    \label{fig: 5}
\end{figure}

For the Implicature task, Mistral$3$-$7$B (KS) and Qwen$3$-$4$B (KS) emerge as the strongest performers, achieving accuracies of $0.76$ and $0.77$, respectively, close to the human baseline of $0.78$. Across models, the “Might be True” class yields the lowest F$1$-scores, indicating limited reasoning depth when handling nuanced implicatures. In contrast to the Agreement task, KS prompting is consistently the most effective strategy for this task, suggesting that external contextual information is particularly beneficial for resolving implicatures. The Presupposition task proves to be the most challenging overall. Notably, Gemma$2$-$2$B (ZS) is an outlier, achieving the highest accuracy of $0.71$, surpassing several larger models such as Mistral and LlaMA. Human performance also declines on this task, with an accuracy of $0.68$. Moreover, several models, including Qwen$2.5$-$7$B and Llama$3.2$-$1$B, fail almost entirely, with accuracies at or below $0.10$. We report that the zero scores arise from limited task understanding under KS due to increased prompt length from example inclusion (prompts given in the Appendix \S \ref{sec: eval}). The results outline that the LLMs exhibit moderate to better performance in our tasks, comparable to human evaluation. Figure \ref{fig: 5} presents the mean accuracy of the models across the three prompt settings as a function of model size. The results indicate that the latest and medium-sized LLMs demonstrate a competitive pragmatic reasoning ability, which can be verified using the models Mistral$3$-$7$B and Qwen$2.5$-$7$B. This is also supported by the performance of Llama$3.2$-$1$B, a recent model with a small parameter size, and MentaLLaMA-$13$B, an older model with a large parameter size.

\begin{figure*}[ht]
    \centering
    \includegraphics[width=1.05\linewidth]{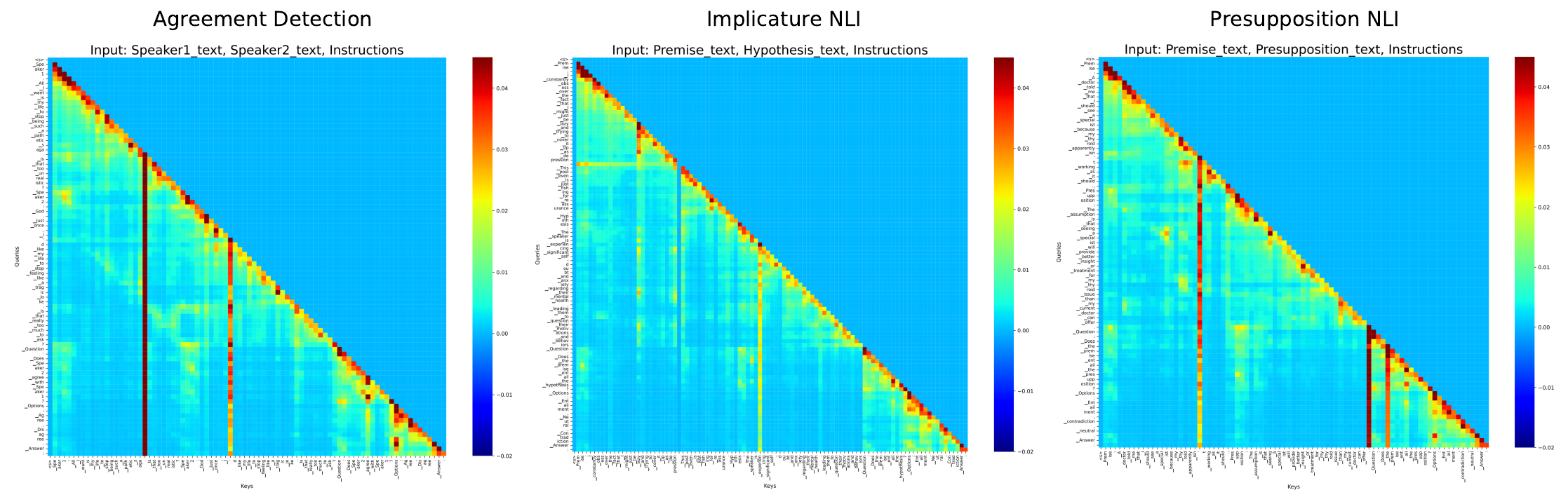} 
    \caption{First-order attention rollout experiment on the MentaLLaMA-7B model for the tasks Agreement Detection, Implicature NLI, and Presupposition NLI from left to right, respectively.} 
    \label{fig: RO}
\end{figure*}

Across all three tasks, human evaluation served as a reference standard, with a few exceptions observed in specific categories. For open-source models and human evaluation, the Presupposition task remains the most challenging reasoning task, followed by the Implicature task and then the Agreement task. These results resemble the performance of LLMs for pragmatics in the open-domain.\footnote{Phi coefficient $(\phi)$ correlations are reported in \S \ref{sec: ARes}.} 

To assess the performance of the models with respect to the label \lq might be true\rq, we further report an error ratio of \texttt{false positives} and \texttt{false negatives}. In particular, we examined the confusion matrix of the classification results to find instances with the incorrect predicted label as \lq might be true\rq in place of \lq true\rq or \lq false\rq. This follows the \texttt{false positives} case for the respective label of \lq might be true\rq. In contrast, as the model predicts \lq true\rq or \lq false\rq in place of \lq might be true\rq, the case falls under \texttt{false negatives}. Following this collection of misclassifications, we evaluate the ratio of \texttt{false positives} to \texttt{false negatives} to find and quantify the error ratio. We calculate this hedging bias, also known as the error asymmetry ratio, and is given by $\frac{FP}{FN}$. We calculate the ratio for Mistral$3$-$7$B as given in table for the two NLI tasks. We also calculate the ratio for all models and present the result in the Appendix \S\ref{sec: ARes}. Mistral$3$-$7$B showed the error ratio close to $1$ in all the prompt setting for the Implicature NLI, which depicts a balanced hedging error distribution. For the Presupposition NLI task ratio slightly greater than $1$, which represents that the model is cautious and hesitant to classify the instances as \lq might be true\rq.

\begin{table}[]
    \centering
    \includegraphics[width=\linewidth]{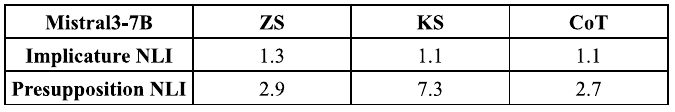} 
    \caption{Error ratio of Mistral$3$-$7$B for the NLI tasks}
    \label{tab: RO}
\end{table}


\noindent \textbf{Analysis with MentaLLaMA}: As the task formatting for fine-tuning the MentaLLaMA model is different as compared to our case, the performance of the model is limited. However, we claim that our proposed tasks are \textit{related but distinct} reasoning tasks and that our \textbf{PRiME} dataset is comparable to the IMHI dataset used by the MentaLLaMA model. 

Figure \ref{tab: RO} represents the first-order attention rollout experiment (as discussed in Section \ref{sec: method}) with MentaLLaMA-$7$B model for the three tasks. \footnote{The axeses consists of the tokens from the input prompts, which are kept in a small font size; we emphasize the demonstration of patterns from the attention maps for all three tasks.} The plots illustrate that the majority of tokens attend to the prior neighboring tokens, and tokens with negligible semantic information, such as the \lq .\rq (dot) token, receive high attention weights. This is mainly due to the softmax function in the attention computation, which dumps the probability mass to a particular token when other tokens do not yield a competitive attention weight \cite{xiao2023smoothquant, bondarenko2023quantizable}. As shown in prior work \citet{liang2025orderlevel}, order-level attention primarily captures syntactic information. Consistent with this, our analysis reveals attention weights concentrating on prior neighboring tokens. However, the resulting attention patterns do not capture token interactions and semantic relationships. We conduct the experiments with other models such as Mistral$3$-$7$B, Mistral$2$-$7$B, LLaMA$3.1$-$8$B, and LLaMA$2$-$7$B to find the similar attention maps as MentaLLaMA-$7$B, these are illustrated in the plots provided in Appendix \S \ref{sec: RO}. Although the intricate semantic relationship could not be captured in our analysis, we present our analysis for the community. We hypothesize that our analysis results stem from the aggregation technique, which averages the attention scores over the head, and then sums up across the layers. In the future, we recommend the community to investigate the initial layers, middle layers, and the later layers separately to find the underlying information flow mechanism inside the LLM architectures.

These findings further highlight our datasets and tasks as a challenging benchmark. On combining the explainability challenges, our work presents valid limitations and challenges of transparency in the architecture of the models. We emphasize the current research gap to the community with respect to open-domain and the mental health domain.

\section{Conclusions}
In this study, we conduct a comprehensive evaluation of pragmatic reasoning in mental health across $12$ LLMs, using the pragmatic phenomena of linguistics. We introduced three datasets and three corresponding tasks that capture diverse dimensions of mental health reasoning. Our results demonstrate that LLMs exhibit moderate to better depth in pragmatic reasoning, which is comparable to the human evaluation. As explainability and transparency of LLMs still remain an ongoing research problem, their deployment in high-stakes mental health settings should be adopted under careful consideration.

\section*{Limitations}
We identify the following limitations of this study:
\begin{enumerate}
    
    \item We propose a simple mapping between mental health and pragmatic reasoning and evaluate the performance of LLMs under this formulation. We highlight that the LLMs exhibit moderate performance to logically relate two pieces of text when framed as a classification task. This can be further engaged with established pragmatics benchmarks and diagnostics frameworks. This study does not aim to develop or support automatic mental health diagnosis, which remains explicitly outside the scope of our work. 
    \item Our analysis using MentaLLaMA attention rollout aims to illustrate how a model fine-tuned on mental health data performs on the proposed tasks and dataset. A deeper investigation into the model’s internal mechanisms or architectural behavior remains an open problem. We consider such analysis as beyond the scope of this work, which primarily focuses on evaluating the performance of current NLP models.
    \item Our analysis is solely based on the LLM-inferencing without any training. The goal in our study is to assess the current progress with the NLP models and discuss their applicability in mental health.
    \item The analysis is conducted with small to medium-sized models with text-based data only. This can be further extended with more LLMs or can be extended to multimodal benchmark and analysis.
\end{enumerate}

\section*{Ethics Statement}
Our work in pragmatic reasoning in mental health addresses valid concerns about individual privacy and ethical considerations. All instances in the study are paraphrased during the cleaning of the \textit{publicly} available CAMS dataset. Furthermore, the datasets used in this study are anonymized before the start of our study, and our research does not involve direct engagement with social media users. For the redefined task, a part of the dataset is generated synthetically and manually curated. Our study is purely observational, specifically based on the capabilities of LLMs. This work does not provide any recommendations for any automatic diagnosis method. 

\bibliography{anthology,custom}

\appendix

\section{Appendix}
\label{sec:appendix}

The model's details used for our evaluation are depicted in Table \ref{tab: 10}.

\begin{table}[ht]
\centering
\begin{tabular}{|c|}
\hline
\textbf{\begin{tabular}[c]{@{}c@{}}Models names\\ Details\end{tabular}}                      \\ \hline
\begin{tabular}[c]{@{}c@{}}Mistral$2$-$7$B\\ mistralai/Mistral-7B-Instruct-v0.2\end{tabular} \\ \hline
\begin{tabular}[c]{@{}c@{}}Mistral$3$-$7$B\\ mistralai/Mistral-7B-Instruct-v0.3\end{tabular} \\ \hline
\begin{tabular}[c]{@{}c@{}}Qwen$3$-$4$B\\ Qwen/Qwen3-4B-Instruct-2507\end{tabular}           \\ \hline
\begin{tabular}[c]{@{}c@{}}Qwen$2.5$-$7$B\\ Qwen/Qwen2.5-7B-Instruct\end{tabular}            \\ \hline
\begin{tabular}[c]{@{}c@{}}Qwen$2.5$-$1.5$B\\ Qwen/Qwen2.5-1.5B-Instruct\end{tabular}        \\ \hline
\begin{tabular}[c]{@{}c@{}}Gemma$2$-$2$B\\ google/gemma-2-2b-it\end{tabular}                 \\ \hline
\begin{tabular}[c]{@{}c@{}}Gemma$3$-$1$B\\ google/gemma-3-1b-it\end{tabular}                 \\ \hline
\begin{tabular}[c]{@{}c@{}}Llama$2$-$7$B\\ llama/llama-2-7b-chat-hf\end{tabular}             \\ \hline
\begin{tabular}[c]{@{}c@{}}Llama$3.1$-$8$B\\ llama/llama-3.1-8B-Instruct\end{tabular}        \\ \hline
\begin{tabular}[c]{@{}c@{}}Llama$3.2$-$1$B\\ llama/Llama-3.2-1B-Instruct\end{tabular}        \\ \hline
\begin{tabular}[c]{@{}c@{}}MentaLLaMA-$7$B\\ klyang/MentaLLaMA-chat-7B\end{tabular}          \\ \hline
\begin{tabular}[c]{@{}c@{}}MentaLLaMA-$13$B\\ klyang/MentaLLaMA-chat-13B\end{tabular}        \\ \hline
\end{tabular}
\caption{Models used for the evaluation of our proposed PRiME benchmark
}
\label{tab: 10}
\end{table}

\subsection{Prompts for data generation} \label{pr_data_ge}

In this section, we present the prompt templates used for the synthetic data generation using \texttt{GPT$5$}. The reasoning parameter is set as \textit{minimal}, and the text verbosity as \textit{low} during the text generation.
\subsubsection{Agreement detection}
Following is the prompt template to generate an agreeing statement by speaker 2:
\begin{lstlisting}
You are a helpful assistant. Given a statement by 'speaker 1', your task is to generate an equivalent statement by 'speaker 2' who agrees with speaker 1. Take this simple example-
Speaker 1: I really can't remember. 2018 marks that I've been pretty solidly depressed for ten years
Speaker 2: I've been depressed for as long as ten years or so, hard to remember.
Now generate an equivalent (concise) statement for 'speaker 2', given the 'speaker 1' for agreeing, 
'Speaker 1': My mom and dad molested me from 0-18 years. My dad got me pregnant at the age of 11. They sold me to other men and allowed them to use me as they pleased for a night.
'Speaker 2':
\end{lstlisting}

Following is the prompt template to generate a disagreeing statement by speaker 2:
\begin{lstlisting}
Given a statement by 'speaker 1', your task is to generate an equivalent statement by 'speaker 2' who disagrees with speaker 1. Take this simple example-
Speaker 1: People who know about my depression, my medication, my self harm stopped talking to me. 
Speaker 2: People who know about my depression, my medication, and my self-harm have been incredibly supportive and continue to reach out to me
Now generate an opposite statement for 'speaker 2', given the 'speaker 1' taking an emotional context- 
'Speaker 1': I am feeling numb right now but cannot help but feel amused about the irony.
'Speaker 2':
\end{lstlisting}

\subsubsection{Implicature NLI}
Following is the prompt template to generate a correct hypothesis, of the premise:
\begin{lstlisting}
Given a premise, your task is to generate a hypothesis about the speaker of the premise. Keep it short
Premise: I'm in serious trouble. Things have completely fallen apart.
Hypothesis:
\end{lstlisting}

Following is the prompt template to generate a false hypothesis, of the premise:
\begin{lstlisting}
Given a premise, your task is to generate an incorrect or opposite hypothesis about the speaker of the premise. Keep it short
Premise: I'm in serious trouble. Things have completely fallen apart.
Incorrect Hypothesis:
\end{lstlisting}

Following is the prompt template to generate a might-be-true hypothesis, of the premise:
\begin{lstlisting}
Given a statement, generate a possible cause or hidden emotion that the speaker might be experiencing. Keep it short
Statement: Everyone is out drinking, smoking weed, blowing off fireworks, and having fun. I'm alone on my computer, as usual. I look at Snapchat, go to stories, and just see my 'friends' having a great time without me. Vodka, weed, all the good things. All I want in my wasteful life is a girlfriend and a good time, but I suppose neither of those will come to me. If 2018 doesn't go well, chuck it. I'm ending it.
Possible cause/hidden emotion:
\end{lstlisting}

\subsubsection{Presupposition NLI}
Following is the prompt template to generate the correct belief of the premise's speaker:
\begin{lstlisting}
Given a statement, your task is to generate the belief or implicit assumption of the speaker. Keep it short
Statement: I need some sort of support or something, I've been screwed by depression for years
Belief: 
\end{lstlisting}

Following is the prompt template to generate the false or misaligned belief of the premise's speaker:
\begin{lstlisting}
Given a statement and belief of the speaker, your task is to generate a false or misaligned belief of the speaker. Keep it short
Statement: I need some sort of support or something, I've been screwed by depression for years
Belief: The speaker believes that they cannot cope with their depression alone and require external help or assistance
False belief:
\end{lstlisting}

Following is the prompt template to generate the uncertain belief of the premise's speaker:
\begin{lstlisting}
Given a statement and belief of the speaker, your task is to generate an 'uncertain' belief of the speaker that could be true or not if given more context/evidence. Keep it short
Statement: I need some sort of support or something, I've been screwed by depression for years
Belief: The speaker believes that they cannot cope with their depression alone and require external help or assistance
Uncertain belief:
\end{lstlisting}


\subsection{Data annotation guideline} \label{anno_gui}

\subsubsection{Annotation guideline for \texttt{Agreement detection}}
\begin{itemize}
    \item Given: Statement by speaker 1, and statement by (independent) speaker 2.
    \item Instructions: Your task will be to annotate by answering to the question: Does speaker 2 agree with speaker 1?
    \item Labels:
    \begin{itemize}
        \item (Agree) If both speakers show the same feelings or emotions.
        \item (Disagree): If both speakers show different or opposite feelings or emotions.
    \end{itemize}
\end{itemize}

Annotate with one of the appropriate labels of (agree) or (disagree). 

\subsubsection{Annotation guideline for \texttt{Implicature NLI}}
\begin{itemize}
    \item Given: A premise and a hypothesis, where the premise will be a social media post, and the hypothesis is an explanation or an underlying cause.
    \item Instructions: Your task will be to annotate by answering to the question: Is the hypothesis definitely true, definitely false, or might be true given the premise?
    \item Labels:
    \begin{itemize}
        \item (Definitely true) If a correct explanation of the cause or intent, or reason behind the premise, is given.
        \item (Definitely false) If the hypothesis is about the opposite emotion of the speaker.
        \item (Might be true) If the hypothesis explains a possible cause or a hidden emotion/psychological state that the speaker of the premise might be experiencing.
    \end{itemize}
\end{itemize}

Annotate with one of the appropriate labels of (definitely true) or (definitely false), or (might be true).

\subsubsection{Annotation guideline for \texttt{Presupposition NLI}}

\begin{itemize}
    \item Given: A premise and its presupposition, where the premise will be a social media post, and the presupposition is an implicit assumption or belief of the premise's speaker.
    \item Instructions: Your task will be to annotate by answering to the question: Is the presupposition definitely true, definitely false, or might be true given the premise?
    \item Labels:
    \begin{itemize}
        \item (Definitely true) If the presupposition accurately captures the author’s belief. 
        \item (Definitely false) If the presupposition reflects a false or misaligned belief.
        \item (Might be true) If the presupposition reflects an ‘uncertain’ belief that could be true or not, if given more context/evidence. 
    \end{itemize}
\end{itemize} 

Annotate with one of the appropriate labels of (Definitely true) or (Definitely false), or (Might be true).

\subsection{Prompts for evaluation} \label{sec: eval}
The following template is used for evaluating the models with the zero-shot prompt setting:
\noindent Task 1: Agreement Detection 
\begin{lstlisting}
You are a mental health support assistant. You are given statements by two independent speakers (speaker 1 and speaker 2). Answer whether speaker 2 agree with speaker 1 or not, by chosing A. agree, B. disagree.
Speaker 1: {speaker\_1\_text}
Speaker 2: {speaker\_2\_text}
Choose the correct option A. agree, B. disagree
Response: [A/B]
\end{lstlisting}

\noindent Task 3: Presupposition NLI
\begin{lstlisting}
You are a mental health support assistant. Given a premise and presupposition, answer whether presupposition is A. definitely true, B. definitely false, or C. might be true given the premise. 
Premise: {premise\_text}
Presupposition: {presupposition\_text}
Choose the correct option A. definitely true, B. definitely false, or C. might be true
Response: [A/B/C]]
\end{lstlisting}

The following template is used for evaluating the models with the k-shot prompt setting:

\noindent Task 1: Agreement Detection
\begin{lstlisting}
You are a mental health support assistant. You are given statements by two independent speakers (speaker 1 and speaker 2). Answer whether speaker 2 agree with speaker 1 or not, by chosing A. agree, B. disagree. Consider following examples:
Speaker 1: I'm done with everything, because as far as i'm concerned, I don't see anything positive anymore.
Speaker 2: I've reached my limit-there's just no light left anywhere I look.
Response: A

Speaker 1: Is there anyone I can talk to who may understand what it is I'm feeling? Are these feelings even real?
Speaker 2: I don't need to talk to anyone about my feelings because I am completely confident in the reality
Response: B

Now choose the correct option A. agree, B. disagree
Speaker 1: {speaker 1 text}
Speaker 2: {speaker 2 text}
Response: [A/B]
\end{lstlisting}

\noindent Task 2: Implicature NLI
\begin{lstlisting}
You are a mental health support assistant. Given a premise and hypothesis, answer whether hypothesis is A. definitely true, B. definitely false, or C. might be true given the premise. Consider following examples:
Premise: If I don't do well I can get kicked off again. I have no friends, nobody to talk to, and don't know how to get it out.
Hypothesis: The speaker is likely experiencing feelings of isolation and anxiety related to their performance, possibly in an academic or social setting (evidently true)
Response: A

Premise: I've been feeling lonely lately and really depressed, and I just want to talk to some people
Hypothesis: The speaker is thriving in his social life, expressing a desire for connection (opposite feeling)
Response: B

Premise: I just need someone to talk to I'm tired of caring and no one else reciprocating. I want a friend. I just want someone to care.
Hypothesis: The speaker is likely experiencing loneliness and a deep yearning for connection and validation (\lq validation\rq is not certainly evident from premise)
Response: C

Now choose the correct option A. definitely true, B. definitely false, or C. might be true
Premise: {premise text}
Hypothesis: {hypothesis text} 
Response: [A/B/C]
\end{lstlisting}

\noindent Task 3: Presupposition NLI
\begin{lstlisting}
You are a mental health support assistant. Given a premise and presupposition, answer whether presupposition is A. definitely true, B. definitely false, or C. might be true given the premise. Consider following examples:
Premise: Looking for some comfort as I've been having a rough time.
Presupposition: The speaker believes that comfort can help alleviate their distress or improve their emotional state (evident belief)
Response: A

Premise: A year ago, depression was at its worst. It was the best thing I ever did, but why am I still suffering a year later? A psychiatrist seems happy that I am doing so much better than I was. 
Presupposition: The speaker believes that their depression has completely disappeared and that they are no longer affected by it at all (opposite belief)
Response: B

Premise: Even though some of my friends tell me otherwise I still feel guilty of what I've done and unworthy of everything
Presupposition: The speaker wonder if their friends are right and if they should feel differently about their guilt and worthiness (uncertain about friends being right, needs more evidence)
Response: C

Now choose the correct option A. definitely true, B. definitely false, or C. might be true
Premise: {premise text}
Presupposition: {presupposition text} 
Response: [A/B/C]
\end{lstlisting}

\noindent The following template is used for evaluating the models with the chain-of-thought prompt setting:

\noindent Task 1: Agreement Detection
\begin{lstlisting}
You are a mental health support assistant. You are given statements by two independent speakers (speaker 1 and speaker 2). Answer whether speaker 2 agree with speaker 1 or not, by chosing A. agree, B. disagree. 
Speaker 1: {speaker 1 text}
Speaker 2: {speaker 2 text} 
Think step-by-step and choose the correct option A. agree, B. disagree
Response: [A/B]
\end{lstlisting}

\noindent Task 2: Implicature NLI
\begin{lstlisting}
You are a mental health support assistant. Given a premise and hypothesis, answer whether hypothesis is A. definitely true, B. definitely false, or C. might be true given the premise. 
Premise: {premise text}
Hypothesis: {hypothesis text} 
Think step-by-step and choose the correct option A. definitely true, B. definitely false, or C. might be true
Response: [A/B/C]
\end{lstlisting}

\noindent Task 3: Presupposition NLI
\begin{lstlisting}
You are a mental health support assistant. Given a premise and presupposition, answer whether presupposition is A. definitely true, B. definitely false, or C. might be true given the premise. 
Premise: {premise text}
Presupposition: {presupposition text} 
Think step-by-step and choose the correct option A. definitely true, B. definitely false, or C. might be true
Response: [A/B/C]
\end{lstlisting}

The following prompt template is presented for the two versions of the MentaLLaMA model. Both models are given the prompt with explicit mention of input text as social media posts, and to generate reasoning as per need, for their better task understanding. The following is the template for zero-shot prompt setting only. For the k-shot and chain-of-thought experiment, the examples and the text \lq think step-by-step\rq are given, respectively, as the prompt template with other models, mentioned above. \\
\noindent Task 1: Agreement Detection
\begin{lstlisting}
Given two posts from two independent authors, answer whether both authors agree or not. Choose the correct option- A. agree, B. disagree. Answer with A or B and do not give any explanations or any other token.
Choose the correct option [A/B]- A. agree, B. disagree
Post 1: {speaker 1 text}
Post 2: {speaker 2 text}  
Response: [A/B]
\end{lstlisting}

\noindent Task 2: Implicature NLI
\begin{lstlisting}
Given a post and its explanation, answer with one of the labels- A. Explanation is 'definitely true' given post, B. Explanation is 'definitely false' given post, C. Explanation is 'might be true' given post. Answer with A, or B, or C, and do not give any explanations or any other token. 
Now choose the correct option [A/B/C]- A. Explanation is 'definitely true' given the post, B. Explanation is 'definitely false' given the post. C. Explanation is 'might be true' given the post
Post : {premise text}
Explanation: {hypothesis text} 
Response: [A/B/C]
\end{lstlisting}

\noindent Task 3: Presupposition NLI
\begin{lstlisting}
Given a post and belief of the author (implicit assumption), check the belief and answer with one of the labels- A. Belief is 'definitely true', B. Belief is 'definitely false', C. Belief is 'might be true' given post. Answer with A, or B, or C, and do not give any explanations or any other token.
Choose the correct option [A/B/C]- A. Belief is 'definitely true', or B. Belief is 'definitely false', or C. Belief is 'might be true'.
Post : {premise text}
Belief: {presupposition text}
Response: [A/B/C]
\end{lstlisting}

The same instructions as given in Appendix \S \ref{anno_gui} are presented to the two Master's students for the human evaluation of our results. The final labels are obtained after resolving disagreement with IAA $\kappa=0.76$. The students are compensated fairly for their contribution.

All the experiments in this work are conducted in the period from 20 November - 30 November 2025.

\subsection{Additional Results} \label{sec: ARes}

\begin{figure*}[ht]
    \centering
    \includegraphics[width=1.05\linewidth]{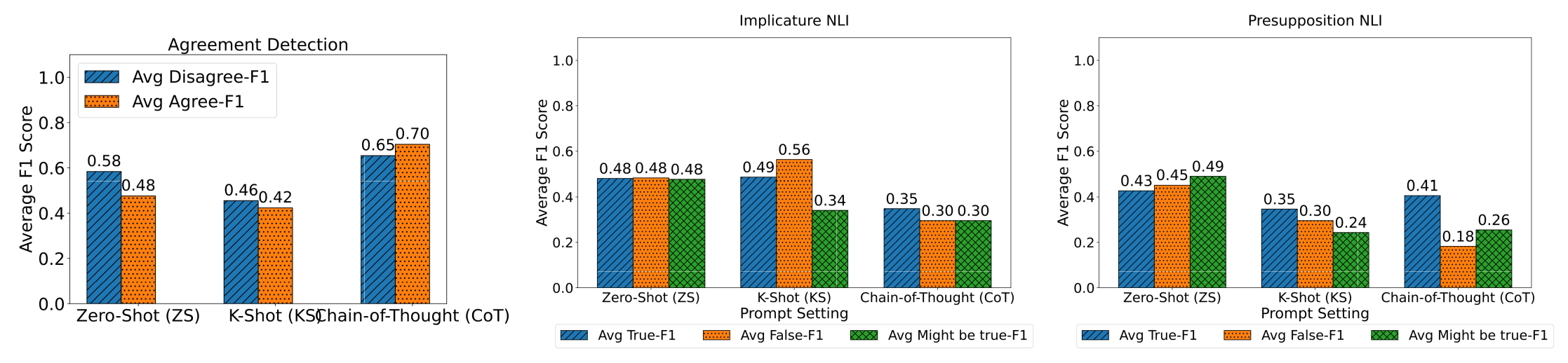} 
    \caption{Average F1-score aggregated for the prompt setting across each label for the three tasks.}
    \label{fig: 4}
\end{figure*}

\noindent \textbf{The Phi correlation coefficients: } The Phi correlation coefficients among the Ground Truth (GT), Human evaluators (Human), and Mistral$3$-$7$B (LLM) are reported in Table \ref{tab: 11}.

\begin{table}[]
\centering
\begin{tabular}{|c|c|c|}
\hline
\textbf{Tasks}                                                          & \textbf{GT-Human} & \textbf{Human-LLM} \\ \hline
\begin{tabular}[c]{@{}c@{}}Task 1:\\ Agreement\end{tabular} & 0.921             & 0.769              \\ \hline
\begin{tabular}[c]{@{}c@{}}Task 2:\\ Implicature\end{tabular}     & 0.695             & 0.541              \\ \hline
\begin{tabular}[c]{@{}c@{}}Task 3:\\ Presupposition\end{tabular}  & 0.601             & 0.486              \\ \hline
\end{tabular}
\caption{Phi coefficient $(\phi)$ correlations among Ground Truth (GT), Human evaluators (Human), and Mistral$3$-$7$B (LLMs).
}
\label{tab: 11}
\end{table}

\begin{table}[]
    \centering
    \includegraphics[width=\linewidth]{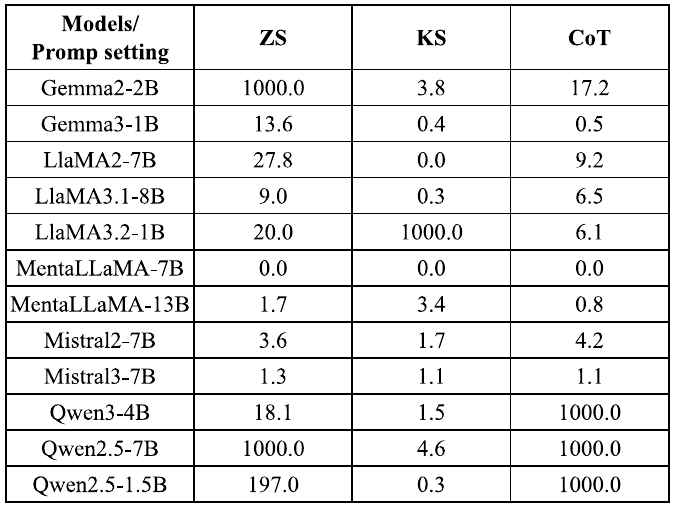} 
    \caption{Error ratio of the models with respect to the label \lq might be true\rq for the Implicature NLI task.}
    \label{tab: 44}
\end{table}

\begin{figure}[h]
    \centering
    \includegraphics[scale=0.28]{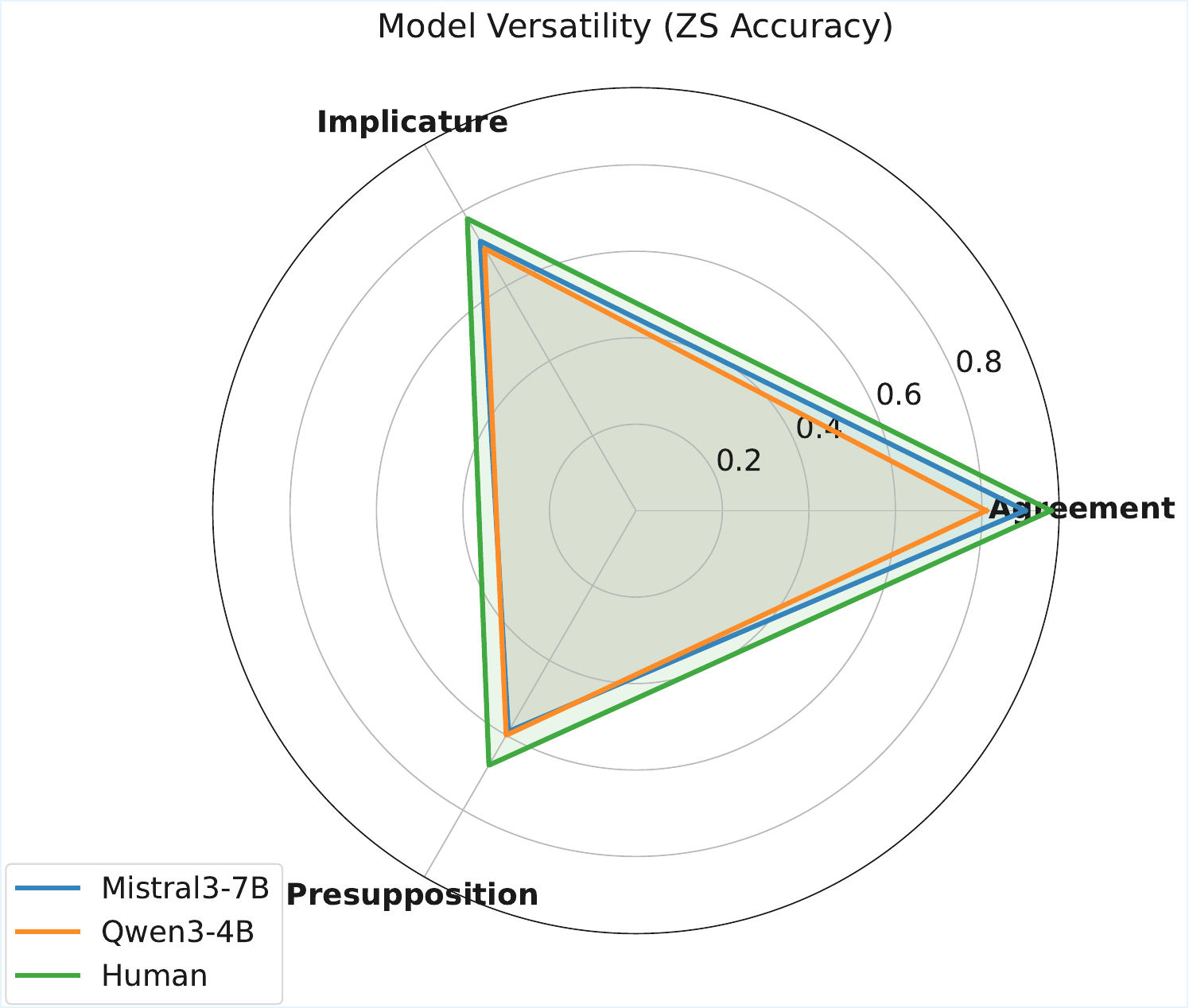} 
    \caption{Versatility of the models on the three tasks based on their performance in the zero-shot setting.}
    \label{fig: 6}
\end{figure}

\noindent \textbf{Versatility of models:} An evaluation of the versatility of the models is conducted using a radar chart, and it is observed that human evaluation forms a large, balanced triangle. It is followed by Mistral$3$-$7$B, achieving a competitive performance compared to all $12$ LLMs, closely followed by Qwen$3$-$4$B. The details are presented in Figure \ref{fig: 6}.

 \begin{table}[]
    \centering
    \includegraphics[width=\linewidth]{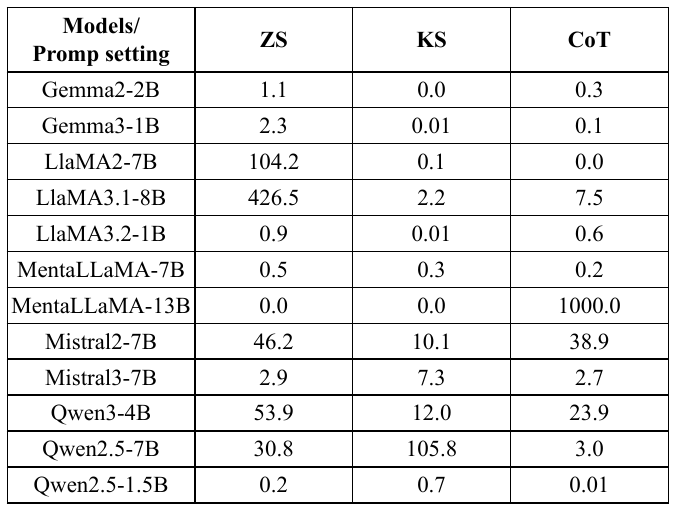} 
    \caption{Error ratio of the models with respect to the label \lq might be true\rq for the Presupposition NLI task.}
    \label{tab: 55}
\end{table}

\begin{figure}[]
    \centering
    \includegraphics[width=1\linewidth]{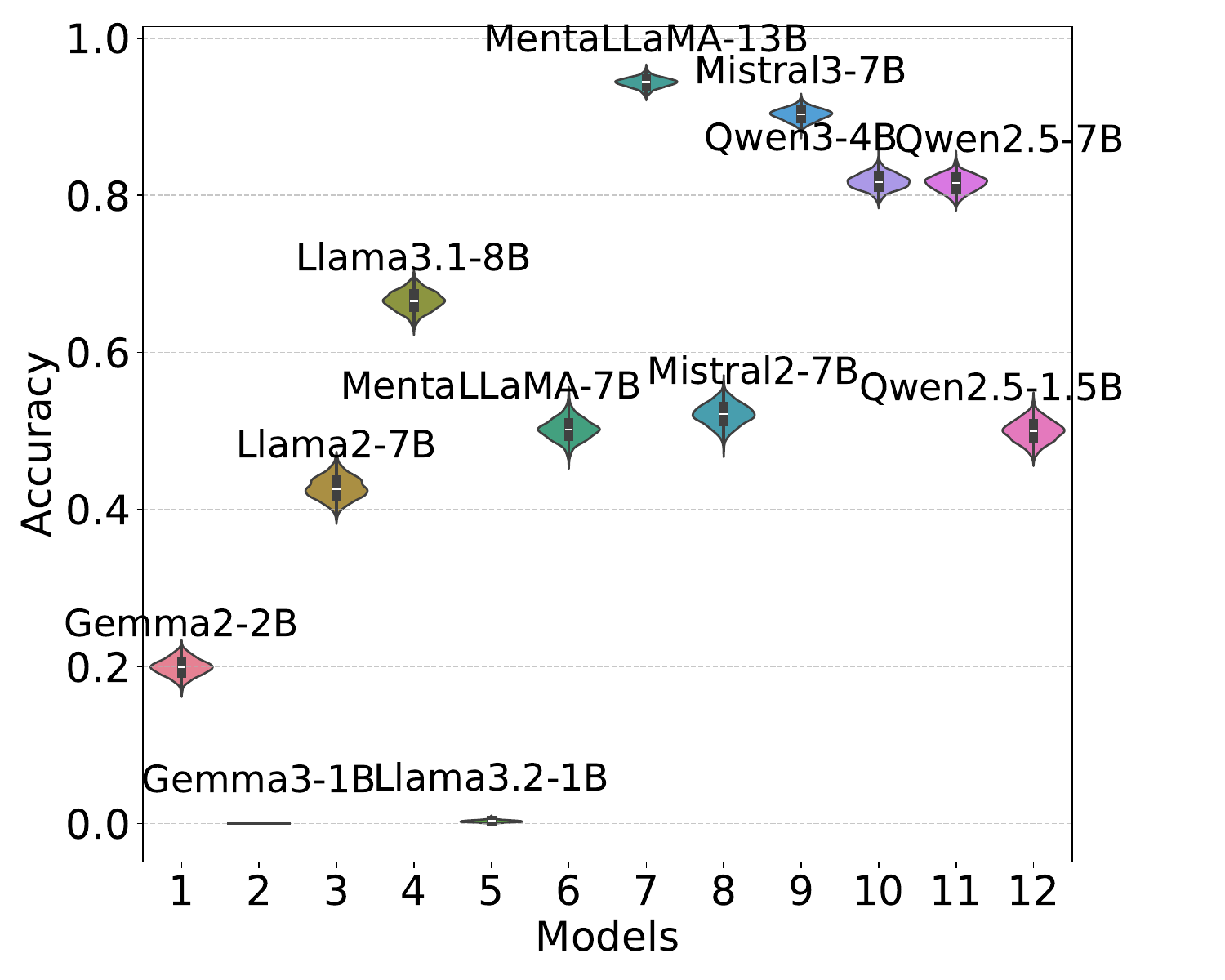} 
    \caption{Bootstrapped accuracies for each model for Agreement Detection in a zero-shot setting.}
    \label{fig: 7}
\end{figure}

\noindent \textbf{Uncertainty Quantification:} To quantify result uncertainty, we compute $95\%$ confidence intervals for the accuracies of all models using bootstrap sampling with $1,000$ iterations. We conducted this analysis for all models, and the resulting confidence intervals showed minimal variability, suggesting stable performance across models. 
The bootstrapped accuracy of all model with 95\% confidence interval is shown for Agreement, Implicature, and Presupposition tasks in Table \ref{tab: 10}, \ref{tab: 16}, and \ref{tab: 12}, respectively. The distribution for the ZS setting on the Agreement task can be found in Figure \ref{fig: 7}.

\begin{table}[]
    \centering
    \includegraphics[width=1\linewidth]{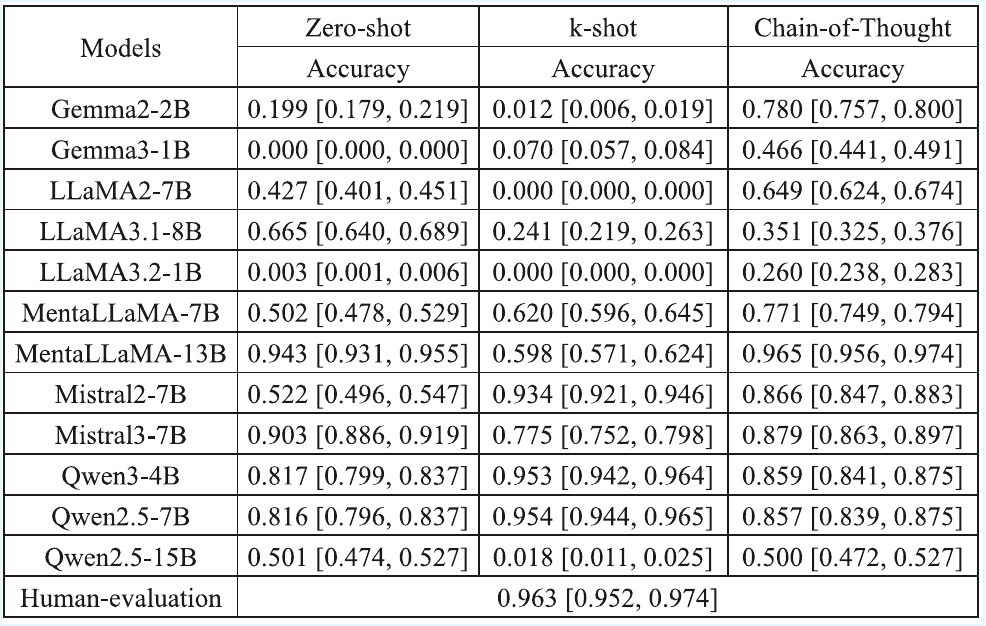} 
    \caption{Bootstrapped accuracy of all models on the Agreement Detection task with a 95\% confidence interval. Values in square brackets indicate the 2.5th and 97.5th percentile accuracies.}
    \label{tab: 10}
\end{table}

\begin{table}[]
    \centering
    \includegraphics[width=1\linewidth]{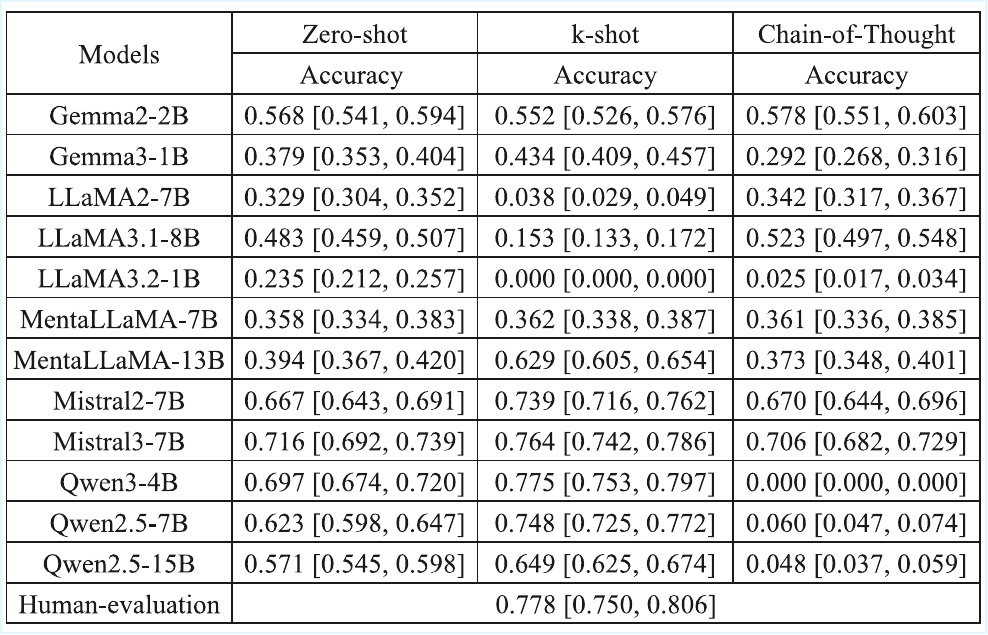} 
    \caption{Bootstrapped accuracy of all models on the Implicature NLI task with a 95\% confidence interval. Values in square brackets indicate the 2.5th and 97.5th percentile accuracies.}
    \label{tab: 16}
\end{table}

\begin{table}[]
    \centering
    \includegraphics[width=1\linewidth]{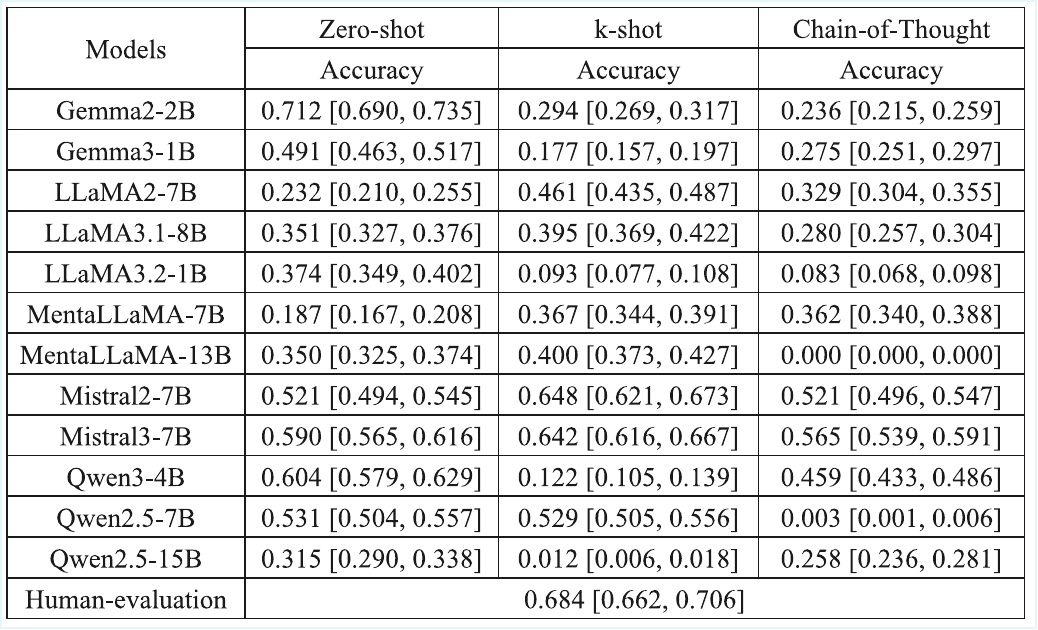} 
    \caption{Bootstrapped accuracy of all models on the Presupposition NLI task with a 95\% confidence interval. Values in square brackets indicate the 2.5th and 97.5th percentile accuracies.}
    \label{tab: 12}
\end{table}

An aggregated average F$1$-score is computed for each label across the three prompt settings for each task. It is observed that the ZS prompt setting established stable F$1$-scores for each of the three tasks. This encourages two further analyses of the LLMs' performance in the ZS setting, details are demonstrated in Figure \ref{fig: 4}. 

\subsubsection{Error (Hedge) Ratio}
Tables \ref{tab: 44} and  \ref{tab: 55} represent the hedging ratio for all models for the Implicature NLI and Presupposition NLI tasks, respectively. For the models where the \texttt{false negatives} was $0$, we substitute an $\epsilon = 0.001$ for a valid ratio $\frac{FP}{FN}$.

\subsubsection{Attention Rollout experiment} \label{sec: RO}

\begin{figure}[]
    \centering
    \includegraphics[width=1\linewidth]{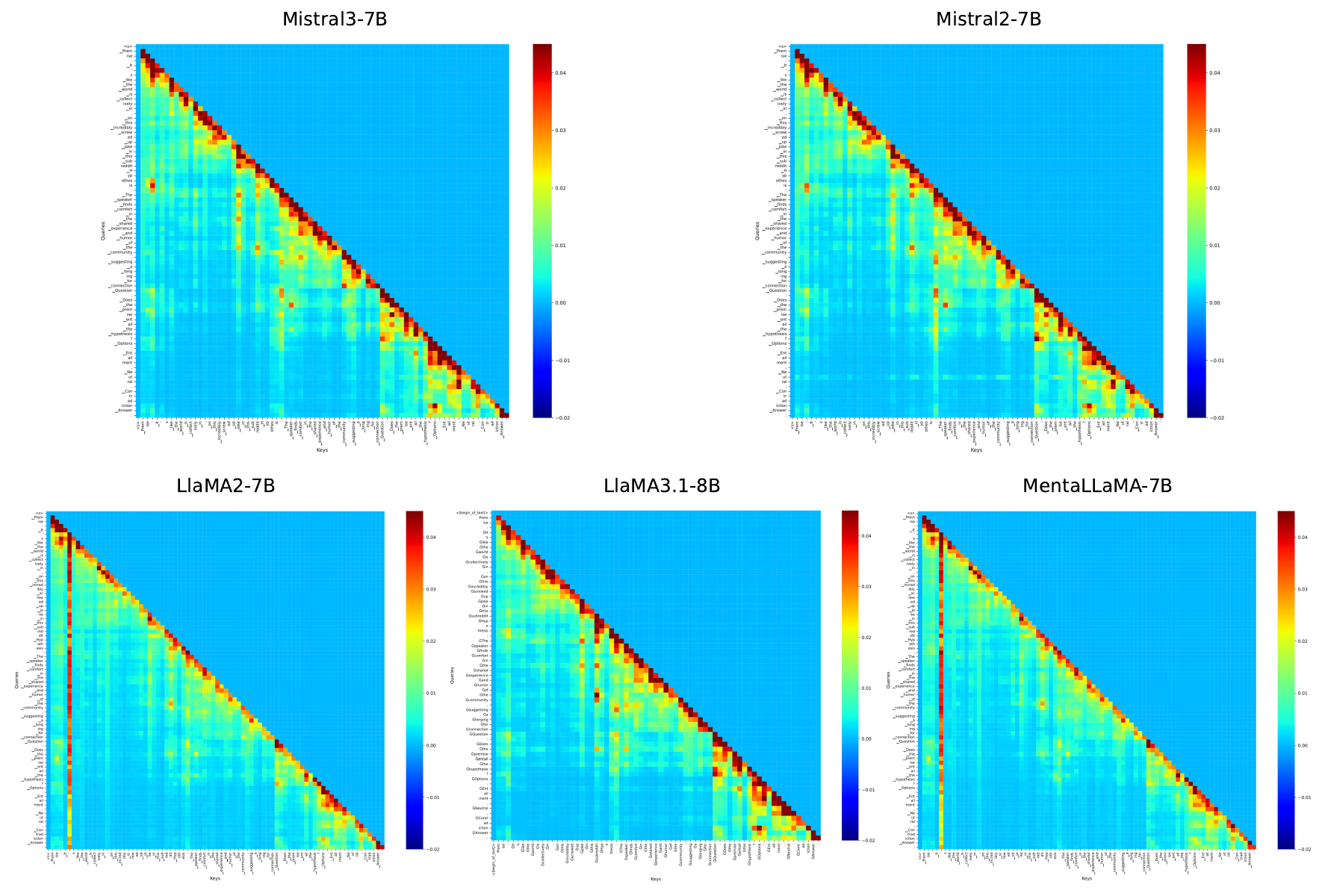} 
    \caption{First-order attention rollout on the premise and hypothesis pair from implicature NLI task with label \lq false\rq.}
    \label{fig: 9}
\end{figure}

\begin{figure}[]
    \centering
    \includegraphics[width=1\linewidth]{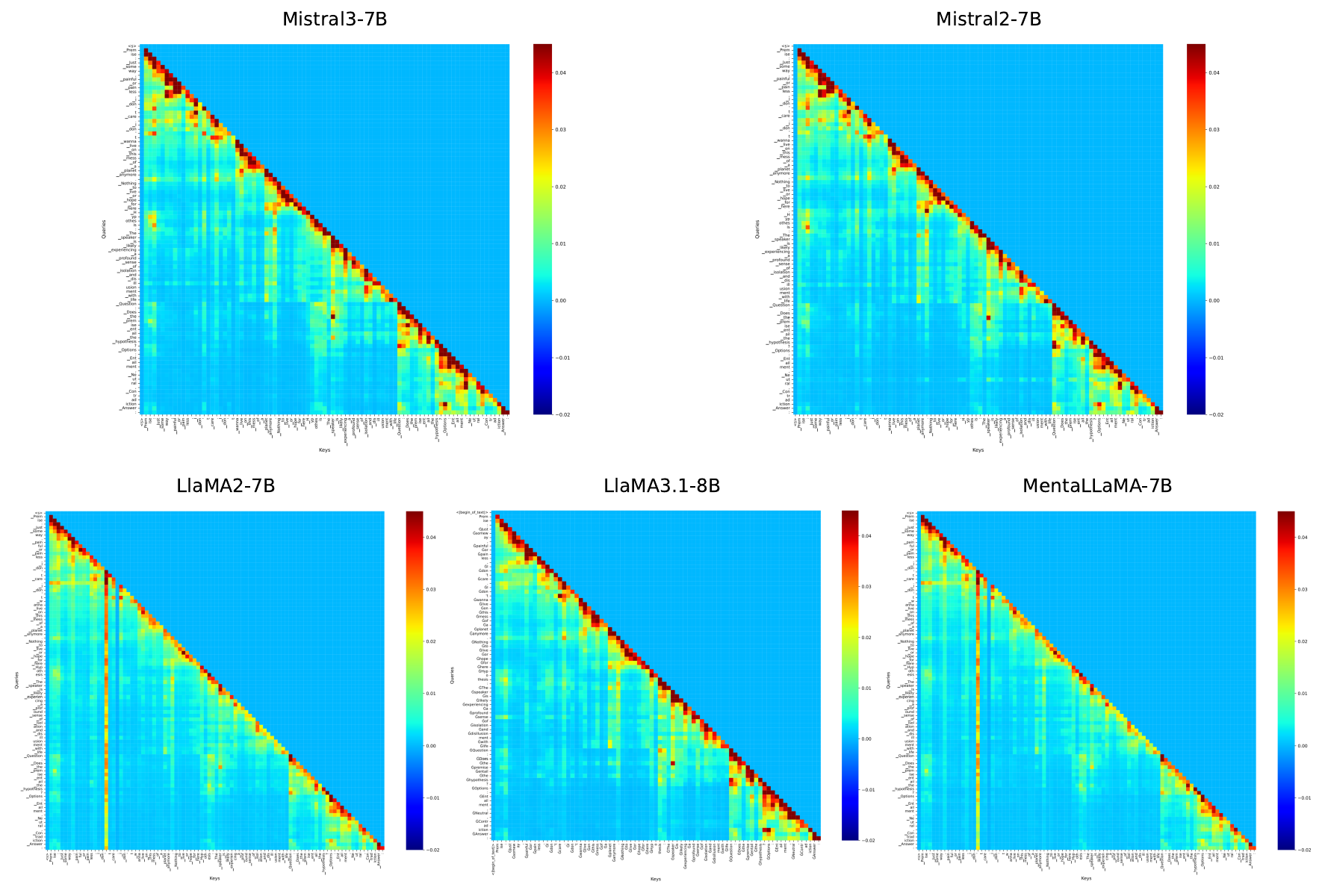} 
    \caption{First-order attention rollout on the premise and hypothesis pair from implicature NLI task with label \lq might be true\rq.}
    \label{fig: 10}
\end{figure}

The plots for the first-order attention rollout experiment for the Implicature NLI task for labels \lq true\rq, \lq false\rq, and \lq might be true\rq are illustrated in Figure \ref{fig: 8}, \ref{fig: 9}, and \ref{fig: 10}, respectively. We highlight that the intricate token relationships and the semantic information from the attention map are an open problem, which we consider a future scope of study. The corresponding prompts for the three plots are:

\noindent \textbf{Prompt for premise-hypothesis pair for label \lq true\rq}:

\begin{figure}[t]
    \centering
    \includegraphics[width=1\linewidth]{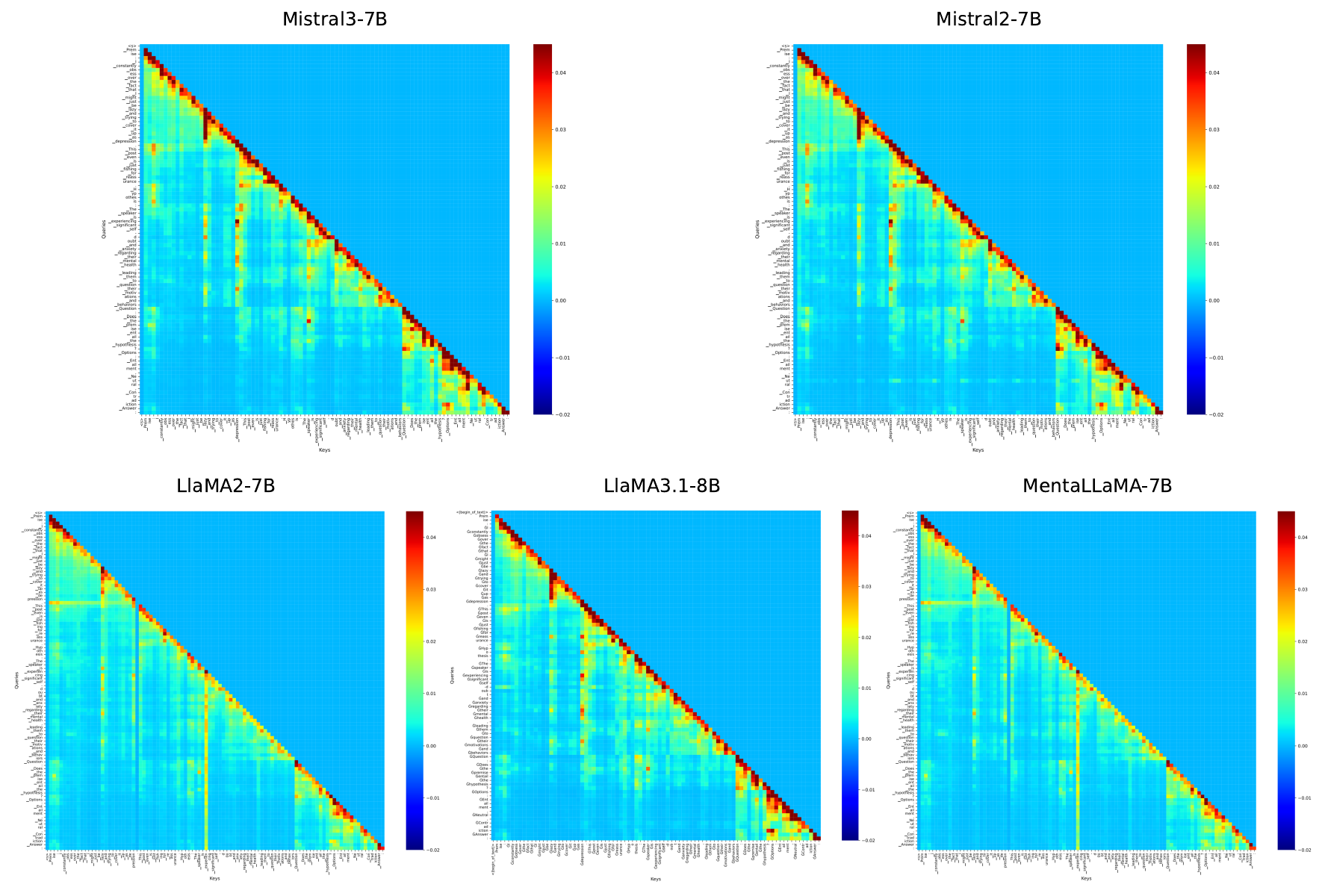} 
    \caption{First-order attention rollout on the premise and hypothesis pair from implicature NLI task with label \lq true\rq.}
    \label{fig: 8}
\end{figure}

\begin{lstlisting}
Premise: {I constantly obsess over the fact that I might just be lazy and trying to cover it up as depression. This post even is just fishing for reassurance}. Hypothesis: {The speaker is experiencing significant self-doubt and anxiety regarding their mental health, leading them to question their motivations and behaviors}. Question: Does the premise entail the hypothesis? Options: Entailment, Neutral, Contradiction Answer:
\end{lstlisting}

\noindent \textbf{Prompt for premise-hypothesis pair for label \lq false\rq}:
\begin{lstlisting}
Premise: {It's like the world is collectively in, on this incredibly screwed up joke in this subreddit}. Hypothesis: {The speaker finds comfort in the shared experience and humor of the community, suggesting a longing for connection}. Question: Does the premise entail the hypothesis? Options: Entailment, Neutral, Contradiction Answer:
\end{lstlisting}

\noindent \textbf{Prompt for premise-hypothesis pair for label \lq might be true\rq}:
\begin{lstlisting}
Premise: {Just somehow, painful or painless, I don't care. I don't wanna live on this mess of a planet anymore. Nothing to live or hope for here.}. Hypothesis: {The speaker is likely experiencing a profound sense of isolation and disillusionment with life}. Question: Does the premise entail the hypothesis? Options: Entailment, Neutral, Contradiction Answer:
\end{lstlisting}

The plots demonstrated are for the models Mistral2-7B, Mistral3-7B, LlaMA2-7B, LlaMA3.1-8B, and MentaLLaMA-7B.

\end{document}